\documentclass[10pt,twocolumn,letterpaper]{article}

\usepackage{iccv}      
\usepackage{multirow}
\usepackage{color}
\usepackage{bbding}
\usepackage{graphicx}
\usepackage[T1]{fontenc}
\definecolor{iccvblue}{rgb}{0.21,0.49,0.74}
\usepackage[pagebackref,breaklinks,colorlinks,allcolors=iccvblue]{hyperref}

\def\paperID{10891} 
\def\confName{ICCV}
\def\confYear{2025}

\title{Robust Dataset Distillation by Matching Adversarial Trajectories}

\author{
Wei Lai$^{1}$, Tianyu Ding$^{2}$, Dongdong Ren$^{1}$,  \\
Lei Wang$^{3}$, 
Jing Huo$^{1}$, Yang Gao$^{1}$, Wenbin Li$^{1}$\thanks{Corresponding author.} \\
\fontsize{10pt}{12pt}\selectfont$^{1}$State Key Laboratory for Novel Software Technology, Nanjing University, China \\
\fontsize{10pt}{12pt}\selectfont $^{2}$Microsoft Corporation, USA \quad $^{3}$University of Wollongong, Australia \\
}
\begin{document}
\maketitle

\begin{abstract}
Dataset distillation synthesizes compact datasets that enable models to achieve performance comparable to training on the original large-scale datasets. However, existing distillation methods overlook the robustness of the model, resulting in models that are vulnerable to adversarial attacks when trained on distilled data. To address this limitation, we introduce the task of ``robust dataset distillation",  a novel paradigm that embeds adversarial robustness into the synthetic datasets during the distillation process. We propose Matching Adversarial Trajectories (MAT), a method that integrates adversarial training into trajectory-based dataset distillation. MAT incorporates adversarial samples during trajectory generation to obtain robust training trajectories, which are then used to guide the distillation process. As experimentally demonstrated, even through natural training on our distilled dataset, models can achieve enhanced adversarial robustness while maintaining competitive accuracy compared to existing distillation methods. Our work highlights robust dataset distillation as a new and important research direction and provides a strong baseline for future research to bridge the gap between efficient training and adversarial robustness.
\end{abstract}

\section{Introduction}
\label{sec:intro}

Deep learning has revolutionized numerous domains, from computer vision to natural language processing \cite{he2016deep, huang2017densely, dosovitskiy2020image, hoefler2021sparsity}. However, the exponential growth in dataset sizes and model complexities has led to demanding training requirements, creating significant barriers to the practical deployment of deep neural networks, especially in resource-constrained environments.
To address these challenges, dataset distillation has emerged as a promising solution, attracting substantial research interest \cite{wang2018dataset, cui2022dc, gou2021knowledge, nguyen2021dataset}. By synthesizing compact datasets that capture essential characteristics of the original data, this technique enables efficient model training while maintaining comparable performance. The reduced volume of training data offers substantial benefits in computational efficiency and resource utilization.

Despite recent advances in dataset distillation \cite{qin2024label, yu2025teddy, huang2024overcoming, deng2024exploiting}, a critical aspect remains unexplored: adversarial robustness. Models trained on distilled datasets inherit vulnerabilities to adversarial attacks, similar to those trained on original datasets \cite{mkadry2017towards, goodfellow2014explaining, ilyas2019adversarial}. These models can drastically misclassify inputs with subtle, imperceptible perturbations. As dataset distillation gains practical adoption, addressing this gap becomes crucial for reliable deployment.

\begin{figure}[!t]
    \centering
    \includegraphics[width=1\linewidth]{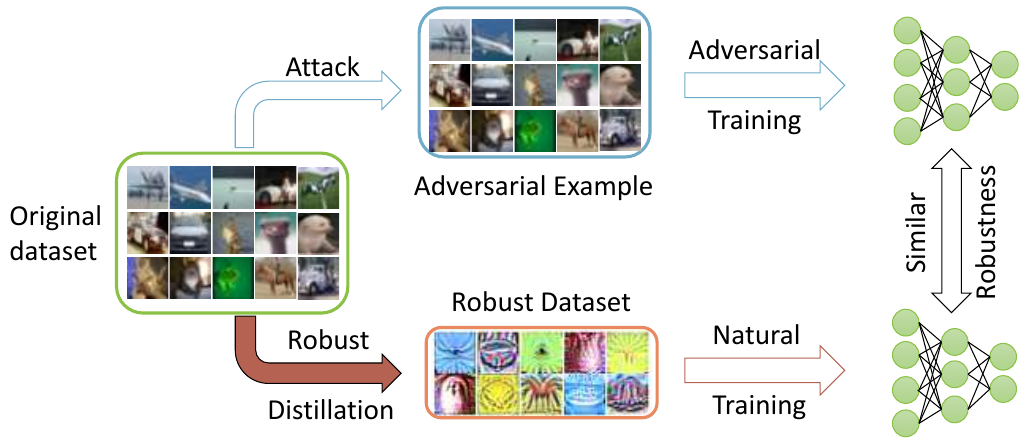}
    \caption{Overview of robust dataset distillation. The process begins with generating adversarial examples from the original dataset. These examples are then used to distill a robust dataset. Models trained on this distilled dataset through standard training naturally acquire adversarial robustness, eliminating the need for explicit adversarial training.}
    \label{fig:overview}
\end{figure}

Traditional defense mechanisms against adversarial attacks have focused primarily on adversarial training \cite{athalye2018obfuscated, uesato2018adversarial}. However, directly applying these methods to distilled datasets presents significant challenges. The core issue lies in the iterative nature of adversarial training, which requires generating adversarial examples during the training process. This iterative approach substantially increases computational costs, which is fundamentally in conflict with the original goal of dataset distillation, which emphasizes efficiency. Consequently, developing methods to achieve robustness through a natural training process on the distilled data is a critical and challenging problem.

\begin{figure}[!t]
    \centering
    \begin{subfigure}[b]{0.49\columnwidth}
        \centering
        \includegraphics[width=\columnwidth]{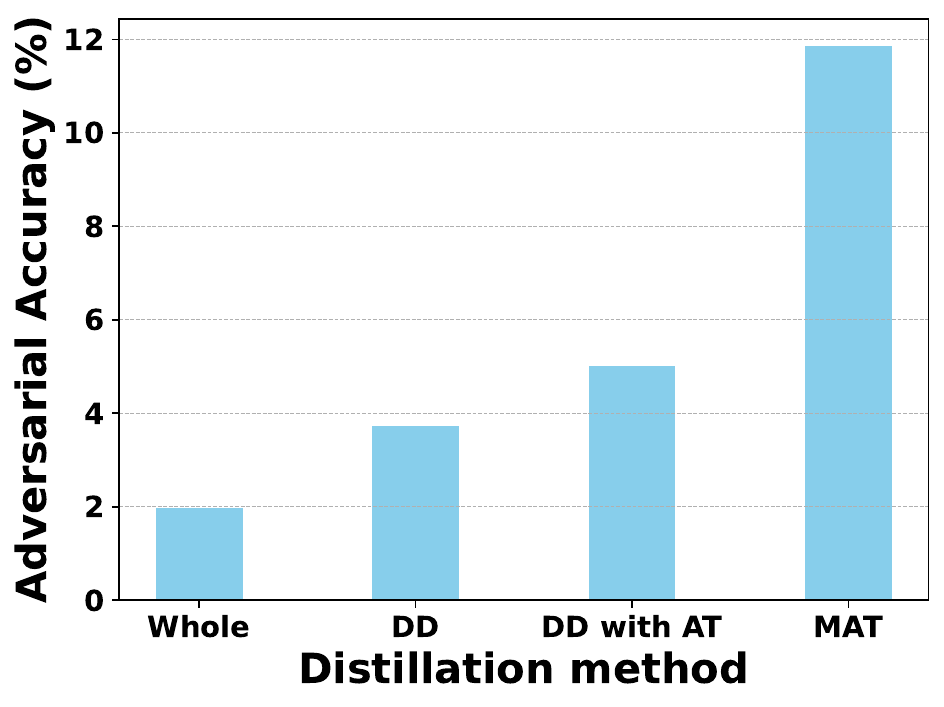}
        \caption{IPC 10}
        \label{fig:adv_10}
    \end{subfigure}
    \begin{subfigure}[b]{0.49\columnwidth}
        \centering
        \includegraphics[width=\columnwidth]{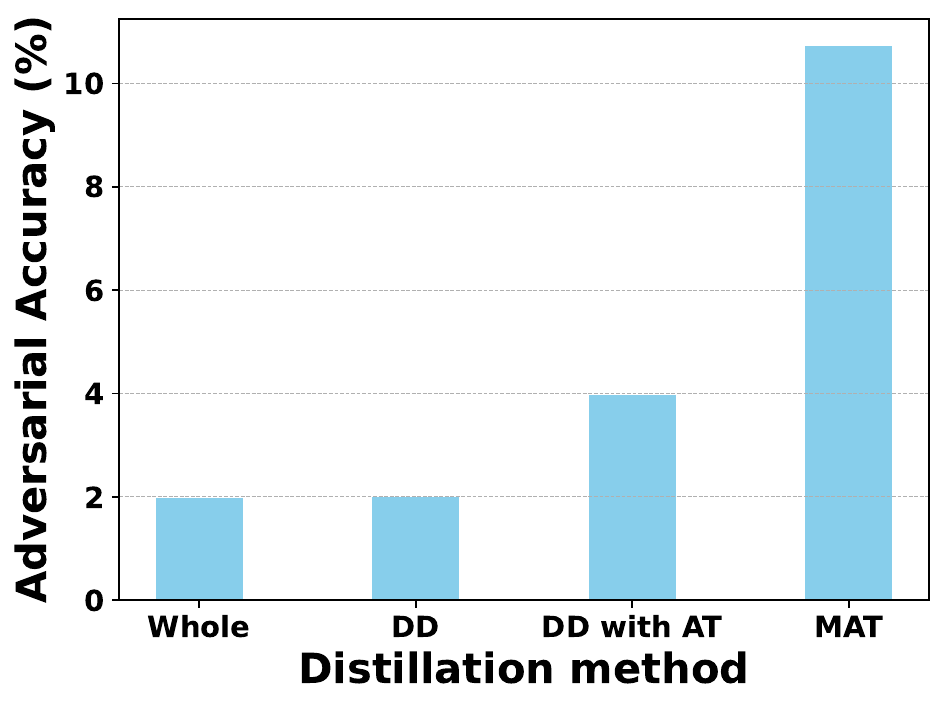}
        \caption{IPC 50}
        \label{fig:adv_50}
    \end{subfigure}
    \caption{Comparison of adversarial accuracy (\%) on CIFAR-10 across different dataset distillation methods with natural training. We set IPC (images per class) to 10 and 50. Adversarial robustness was evaluated using PGD-10 attacks \cite{mkadry2017towards} with $\epsilon$ = 4/255. As a comparison, we include the adversarial accuracy of a naturally trained model on the full dataset. DD refers to standard Dataset Distillation using MTT \cite{cazenavette2022dataset}, while DD with AT uses unmodified adversarial training trajectories, and MAT (our method) employs smoothed adversarial training trajectories.}
    \label{fig:r}
\end{figure}

To address this challenge, we focus on \textit{robust dataset distillation}, a novel paradigm that incorporates adversarial robustness directly into the distillation process. This approach enables models to acquire robustness through natural training on the distilled dataset, eliminating the need for expensive adversarial training, as illustrated in~\cref{fig:overview}. A straightforward approach to this problem involves adapting trajectory-matching-based dataset distillation to work with adversarial trajectories. Traditional trajectory-matching dataset distillation methods align model weight variations on distilled data with those observed in the original dataset. However, our experiments reveal that simply replacing standard trajectories with adversarial ones not only fails to improve robustness but also degrades accuracy on clean data. We attribute this failure to the challenging nature of fitting rapidly changing model weights during adversarial training. In response, we propose smoothing the adversarial training trajectories to facilitate better fitting. By reducing abrupt changes in the model's parameter space, this smoothing strategy helps the model adjust more gradually to adversarial perturbations, promoting stable convergence and improved robustness. Our method, MAT (Matched Adversarial Trajectories), significantly enhances robustness while maintaining competitive accuracy, as shown in~\cref{fig:r}. This work represents a solution to the robust dataset distillation problem. Extensive experiments across multiple datasets validate our approach's effectiveness. Furthermore, we demonstrate the versatility of our method by successfully integrating it with various adversarial training approaches, including PGD-AT \cite{mkadry2017towards}, TRADES \cite{zhang2019theoretically}, and MART \cite{wang2019improving}.

In summary, we make the following key contributions:
\begin{itemize}
    \item We emphasize the importance of robust dataset distillation,  as most existing dataset distillation methods prioritize accuracy and overlook robustness, our research underscores the necessity of embedding robustness directly into the compressed dataset. This enables a natural training process that inherently supports adversarial resilience.
    \item We identify a fundamental limitation in directly employing adversarial trajectory matching to resolve the robust dataset distillation task: the rapid weight changes during adversarial training prevent the effective embedding of robust features into compressed datasets. Furthermore, we propose addressing this issue by smoothing the training trajectory using an EMA mechanism.
    \item We develop the \textit{Matching Adversarial Trajectory (MAT)} method, which leverages smoothed adversarial training trajectories to create distilled datasets with inherent robustness properties. As experimentally verified, our approach enables models to acquire adversarial robustness through natural training on the distilled data.
\end{itemize}

\section{Related Work}
\label{sec:formatting}


\subsection{Dataset distillation}
Dataset distillation, also known as dataset compression, has garnered significant attention as a means of reducing the size of training datasets while preserving their informational content. Early approaches primarily focused on core set selection \cite{farahani2009facility, sener2017active, welling2009herding}, which involves selecting representative samples from the original dataset for retention. However, these methods often suffer from performance limitations and scalability issues when applied to large-scale image datasets. 

Gradient matching-based methods \cite{zhao2020dataset} optimize the synthetic dataset by minimizing the discrepancy between the gradients obtained from training models on the synthetic and original data. Zhao \textit{et al.} \cite{zhao2021dataset} further innovated on this approach by introducing differentiable Siamese augmentation within gradient matching, thereby enhancing the effectiveness of dataset distillation. Feature distribution matching-based methods \cite{zhao2023dataset} have been developed to significantly lower the computational costs associated with dataset distillation. Zhao \textit{et al.} \cite{zhao2023improved} further refined these methods to enhance their performance. Representative works in this category include CAFE \cite{wang2022cafe}, which matches features across various network layers, and DataDAM \cite{sajedi2023datadam}, which focuses on matching feature attention maps. 

Trajectory matching-based methods~\cite{cazenavette2022dataset} optimize the synthetic dataset by aligning the parameter update trajectories of models trained on the original and synthetic data. Building on this, Du \textit{et al.} \cite{du2023minimizing} reduced the error in the distilled dataset by minimizing the accumulated trajectory, while Liu \textit{et al.} \cite{liu2024dataset} proposed an automated approach to match the closest synthetic data trajectories to the original data trajectories, improving the performance of the distilled data. 
These advancements collectively illustrate the progressive efforts in the field of dataset distillation to enhance efficiency, scalability, and performance. However, despite these advancements, no prior work has addressed adversarial robustness in distilled datasets, underscoring the significance of our approach in tackling this critical gap.

\subsection{Adversarial defense}

Adversarial defenses aim to enhance model robustness against adversarial attacks. One of the most well-established training methods in this field is adversarial training \cite{mkadry2017towards, goodfellow2014explaining}, which involves generating adversarial examples during training and incorporating them into the learning process. Two classical approaches include a Fast Gradient Sign Method (FGSM) adversarial training \cite{goodfellow2014explaining}, which employs single-step gradient ascent, and Projected Gradient Descent (PGD) adversarial training \cite{mkadry2017towards}, which extends gradient-based attacks to multiple steps. In addition to these foundational methods, TRADES~\cite{zhang2019theoretically} by~Zhang ~\textit{et~al.} investigates the trade-off between model accuracy and robustness, while \cite{wang2019improving} introduces an error classification-aware strategy to improve overall robustness. Wu \textit{et al.} \cite{wu2020adversarial} further enhance adversarial training by considering perturbations in model weights.  Besides, studies have shown that models pre-trained on large datasets exhibit stronger resistance to adversarial attacks \cite{hendrycks2019using}, and data augmentation \cite{rebuffi2021fixing} contributes to robustness by generating diverse adversarial samples to enrich training data.

Several studies have also explored robust dataset distillation. For instance, Xue \textit{et al.}~\cite{xue2024towards} incorporate curvature regularization in the loss function to improve dataset robustness, while Wu \textit{et al.}~\cite{wu2022towards} utilize meta-learning to achieve a robust distilled  dataset. However, they have notable limitations: Wu \textit{et al.} generate a dataset comparable in size to the original, reducing the efficiency gains of distillation, while Xue \textit{et al.} only evaluate robustness under relatively weak adversarial attacks. In contrast, our method takes a fundamentally different approach by optimizing the training trajectories of the teacher model, embedding robustness into the learning process from the outset. Moreover, our approach addresses both efficiency and robustness concerns by producing a compact distilled dataset that significantly reduces computational overhead while maintaining strong resilience under more challenging adversarial conditions.

\section{The proposed method}

Robust dataset distillation aims to create a compact yet robust dataset \( \mathcal{S} \) from an original dataset \( \mathcal{D} \). The key objective is to enable models trained on \( \mathcal{S} \) to achieve performance comparable to those trained on \( \mathcal{D} \) and, at the same time,  adversarial robustness, all without requiring computationally expensive adversarial training. In this section, we present our method based on matching a smooth robust trajectory. Note that while our approach shares similarities with some existing methods by pre-training a model and preserving its training trajectories across multiple matching iterations, it has a crucial distinction: it needs to handle adversarial training trajectories rather than conventional ones.

\subsection{Preliminaries}

\textbf{Adversarial training.} As a fundamental approach to enhance the robustness of the model against adversarial attacks, adversarial training generates adversarial examples through white-box attack techniques and trains models on these examples. This process can be formalized as,
\begin{equation}\small
\min_{\theta} \mathbb{E}_{(x, y) \sim \mathcal{D}} \left[ \max_{\|x' - x\| \leq \epsilon} \mathcal{L}\left(f_{\theta}(x'), y\right) \right], 
\end{equation}
where \( x' \) denotes the adversarial example, \( \theta \) represents the model parameters, \( \mathcal{D} \) is the original dataset, \( (x, y) \) is an input sample and its corresponding label, \( \epsilon \) is the maximum allowed norm of the adversarial perturbation, \( \mathcal{L} \) is the loss function, and \( f_{\theta} \) is the model function parameterized by \( \theta \).

Adversarial training manifests itself as a bi-level optimization problem: each iteration requires gradient ascent for the inner maximization followed by gradient descent for the outer minimization. This dual optimization structure makes the training process computationally intensive.

\noindent\textbf{Robust dataset.}  A robust dataset embeds adversarial robustness directly into its structure, allowing models to achieve robustness through standard gradient descent training alone. This approach significantly reduces computational overhead while maintaining defensive capabilities. We define the optimization objective as,
\begin{equation}\small
\begin{aligned}
\min_{\mathcal{S} } \quad & \mathbb{E}_{(x, y) \sim \mathcal{D}_{\mathrm{Test}}} \left[ \max_{\|x' - x\| \leq \epsilon} \mathcal{L}\left(f_{\theta^{\mathcal{S}}}(x'), y\right) \right], \\
\text{s.t.} \quad & \theta^{\mathcal{S}} = \arg\min_{\theta} \mathcal{L}_{\mathcal{S}}(\theta),\label{eq:robust dataset}
\end{aligned}
\end{equation}
where $\mathcal{S}$ is the robust dataset, and $\theta^{\mathcal{S}}$ denotes the model parameters trained on $\mathcal{S}$. 
This formulation encapsulates our goal: creating a dataset with inherent robustness properties that enables efficient training while maintaining strong adversarial defenses. By embedding robustness directly into the data, we can streamline the training process while preserving the model's resistance to adversarial attacks.

\subsection{Matching training trajectories}


The trajectory matching-based dataset distillation process begins with training a model on the complete original dataset and recording the model parameters at each epoch's conclusion. These sequential parameter snapshots form an expert trajectory to guide the subsequent distillation process.

More specifically, during each distillation iteration, we randomly select a timestamp $t$ that is smaller than the preset maximum timestamp $T$ and retrieve the corresponding model weight $\theta^{\mathcal{D}}_t$ as a starting student model. The weight parameter of the expert trajectory at timestamp $t+M$, \textit{i.e.}, $\theta^\mathcal{D}_{t+M}$, serves as the target weight. Meanwhile, we train the student model on the distilled dataset for $N$ gradient descent steps to obtain $\theta^\mathcal{S}_{t+N}$. After training the student model, we optimize the alignment between the parameters of the student model $\theta^\mathcal{S}_{t+N}$ and the target parameters $\theta^\mathcal{D}_{t+M}$. The discrepancy between the student and teacher trajectories is defined by the difference in model weights, normalized by the weight change magnitude $\Delta \theta = \|\theta^\mathcal{D}_{t+M} - \theta^\mathcal{D}_t\|$. The objective of the matching process can be formulated as,
\begin{equation}\small
L_{\text{match}} = \frac{\|\theta^\mathcal{S}_{t+N} - \theta^\mathcal{D}_{t+M}\|^2}{\|\theta^\mathcal{D}_{t+M} - \theta^\mathcal{D}_{t}\|^2}.
\end{equation}
The general goal of trajectory matching is to minimize the aforementioned objective function, ensuring that the parameters of the student model closely track the changes in the teacher’s trajectory. Building  upon the original method, automatic trajectories-matching dataset distillation (ATT)~\cite{liu2024dataset} introduces a strategy for automatically selecting the optimal parameter $N$, expressed as,
\begin{equation}\small
N = \arg\min_{N} \|\theta^\mathcal{S}_{t+N} - \theta^\mathcal{D}_{t+M}\|^2.
\end{equation}
This formulation enables the automatic selection of student model weights that exhibit the closest alignment with the target weights during the trajectory matching process. This approach eliminates the potential errors and biases associated with manually setting fixed parameters. Our methodology incorporates this automated selection strategy.

\begin{figure}[!t]
    \centering
    \includegraphics[width=0.78\columnwidth]{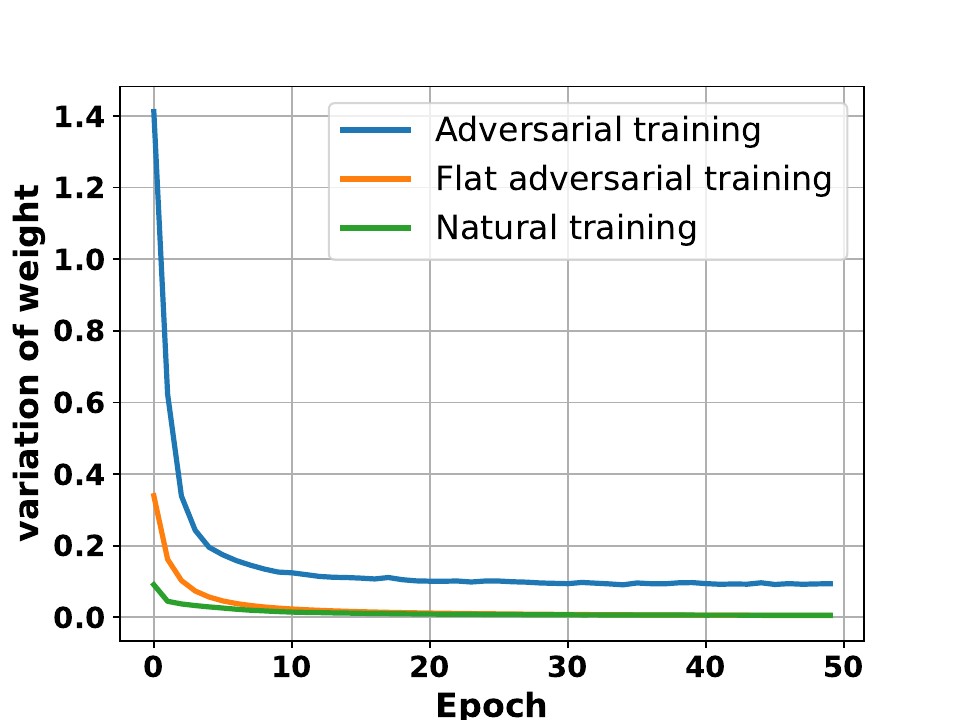}
    \caption{The variance of weight during training. We compared the weight variation during training across three conditions: natural training, adversarial training, and adversarial training in our MAT (flat adversarial training). }
    \label{fig:loss}
\end{figure}

\begin{figure}[!t]
    \centering
    \includegraphics[width=0.9\linewidth]{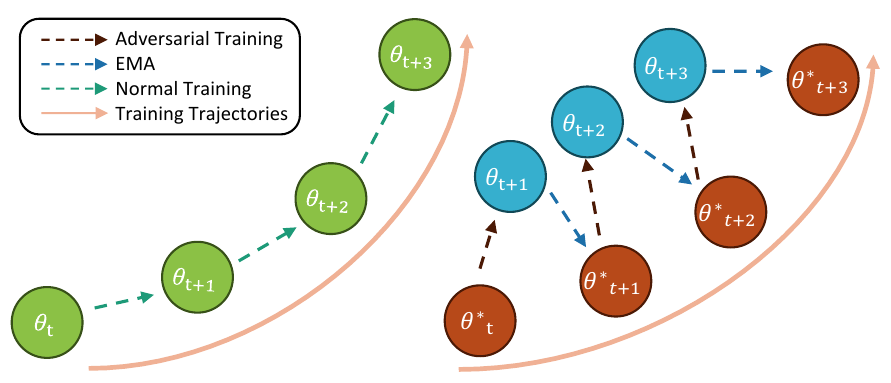}
    \caption{Workflow of trajectory generation in our method. Unlike previous approaches illustrated on the left, our method enhances model robustness via adversarial training and applies EMA to smooth the model's training trajectories.}
    \label{fig:method}
\end{figure}

A straightforward approach to achieving the robustness of a distilled dataset appears to be simply substituting conventional training with adversarial training. This method is expected to incorporate adversarial examples during the training process to enhance the model's resilience against input perturbations. Ideally, by utilizing the robust training trajectory as a reference, we could align the training trajectory of the distilled dataset with that of adversarial training. Then we would, as thought, achieve the goal to enable the model to achieve adversarial robustness through standard training on the distilled dataset alone. However, our experiments revealed an unexpected outcome: rather than improving robustness, the above approach led to a significant degradation in model accuracy.

\subsection{Generating adversarial buffer}
Through a detailed investigation of this phenomenon, we identified that the root cause lies in the rapid fluctuations of model weights during adversarial training, resulting in substantial parameter disparities across different epochs along the trajectory. These dramatic weight variations complicate the precise matching of trajectories during the distillation process, ultimately compromising performance. To validate this hypothesis, we conducted a visual analysis of the weight variant, as plotted in Fig.~\ref{fig:loss}. The results confirmed that adversarial training induces more significant weight variation than natural training, validating our assumption.

To address these rapid weight variations, we propose a trajectory-smoothing approach using the well-established Exponential Moving Average (EMA) algorithm. EMA mitigates weight fluctuations by computing a weighted average of weights from previous and current iterations, effectively smoothing the training trajectory. The original training trajectory follows the standard adversarial training update rule, as depicted in~\cref{fig:method}.

With the incorporation of EMA, our modified update rule is formulated as,
\begin{equation} \small
\begin{aligned}
& \theta^{\text{EMA}}_t = \alpha \theta^{\text{EMA}}_{t-1} + (1 - \alpha) \theta_t, \\
\text{s.t.} \quad & \theta_{t} = \theta_{t-1} - \eta \nabla_{\theta} \mathcal{L}\left(\theta_{t-1}, x', y\right),
\end{aligned}
\end{equation}
where \( \alpha \) is the smoothing coefficient, \( \theta^{\text{EMA}}_t \) represents the EMA-smoothed model weights, $\eta$ is the learning rate, \( \theta_t \) denotes the model weights at iteration \( t \), \( \nabla \mathcal{L} \) is the gradient of the loss function, \( x' \) is the adversarial example, and \( y \) is the label. 
By incorporating EMA, each training step no longer solely depends on the immediate gradient update $\theta_{t}$ but instead blends it with past model states to produce $\theta^{\text{EMA}}_t$. This mechanism effectively mitigates high-frequency oscillations in the parameter space, especially under adversarial gradients. As a result, training becomes more stable, and the parameter trajectory is less susceptible to abrupt shifts, which can be particularly beneficial in adversarial settings where the loss landscape fluctuates rapidly. Moreover, smoothing the parameter updates helps the model converge to flatter minimizers, improving both generalization and robustness. Our experiments demonstrate that this smoothed trajectory not only reduces weight divergence but also improves the reliability of the distillation process, as evidenced by the diminishing gap between adversarial and natural weight updates in later training stages (Fig.~\ref{fig:loss}).

It is important to note that EMA is just one of many possible smoothing techniques. Other methods, such as Stochastic Weight Averaging (SWA)~\cite{izmailov2018averaging}, can also be incorporated into our approach to achieve similar trajectory smoothing benefits.

\section{Experiment}

\begin{table*}[ht]
\large
\centering
\caption{Performance comparison of standard and adversarial accuracy (\%) on CIFAR-10 across different methods with varying IPC (Images Per Class) values. We used PGD-10 with $\epsilon$ = 4/255 to test models' robustness. We reproduced all baseline results in the adversarial defense scenario. The \textbf{best} results in each column are highlighted for clarity.}
{%
\small
\begin{tabular}{l c c c c c c}
    \toprule
    IPC & \multicolumn{2}{c}{1} & \multicolumn{2}{c}{10} & \multicolumn{2}{c}{50} \\
    \cmidrule(lr){2-3} \cmidrule(lr){4-5} \cmidrule(lr){6-7}
    Performance & Standard acc & Adversarial acc & Standard acc & Adversarial acc & Standard acc & Adversarial acc \\
    \midrule
    DC~\cite{zhao2020dataset} & 28.03$\pm$0.52 & 8.98$\pm$0.57 & 43.23$\pm$0.54 & 5.18$\pm$0.83 & 53.18$\pm$0.50 & 3.70$\pm$0.09 \\
    DM~\cite{zhao2023dataset} & 24.17$\pm$0.22 & 11.77$\pm$0.31 & 41.53$\pm$1.08 & 3.71$\pm$0.66 & 58.65$\pm$0.15 & 4.24$\pm$0.42 \\
    DSA~\cite{zhao2021dataset} & 26.82$\pm$0.25 & 10.14$\pm$0.27 & 43.44$\pm$0.63 & 8.66$\pm$0.69 & 56.33$\pm$0.42 & 4.27$\pm$0.32 \\ 
    MTT~\cite{cazenavette2022dataset} & \textbf{38.61$\pm$1.01} & 6.48$\pm$ 0.94 & 53.99$\pm$1.64 & 3.72$\pm$0.43 & 62.73$\pm$1.13 & 1.99$\pm$0.27 \\
    FTD~\cite{du2023minimizing} & 36.48$\pm$1.00 & 6.41$\pm$1.06 & 45.16$\pm$1.18 & 3.23$\pm$0.53 & 63.07$\pm$0.43 & 2.43$\pm$0.18 \\
    ATT~\cite{liu2024dataset} & 37.30$\pm$0.69 & 5.58$\pm$1.13 & \textbf{55.28$\pm$0.87} & 5.55$\pm$0.39 & \textbf{64.35$\pm$0.37} & 2.87$\pm$0.57 \\
    \midrule
    MAT (Ours) & 35.96$\pm$0.74 & \textbf{14.33$\pm$0.63} & 53.04$\pm$0.82 & \textbf{11.85$\pm$0.80} & 62.86$\pm$0.76  & \textbf{10.72$\pm$0.96} \\
    \bottomrule
\end{tabular}%
}
\label{tab:cifar10}
\end{table*}

\subsection{Experiment settings}

\textbf{Datasets.} We conducted our experiments on three datasets: CIFAR-10, CIFAR-100, and Tiny ImageNet.
\begin{itemize}
    \item CIFAR-10 and CIFAR-100~\cite{krizhevsky2009learning}: Popular image classification datasets, each containing 60,000 32$\times$32 color images. CIFAR-10 has 10 classes, while CIFAR-100 has 100 classes.
    \item Tiny ImageNet~\cite{deng2009imagenet}: This dataset includes 100,000 64$\times$64 color images distributed over 200 classes, with 500 training images and 50 validation images per class. 
\end{itemize}

\noindent\textbf{Evaluation details and Baselines.} Our experimental procedure follows methodologies established in prior research~\cite{cazenavette2022dataset, du2023minimizing, liu2024dataset} and consists of two steps: distillation and evaluation. In the distillation step, we first generate a robust dataset using our MAT algorithm. By default, we employ PGD-AT \cite{mkadry2017towards} with $\epsilon$ = 4/255 for adversarial training to obtain trajectories, then generate a distilled dataset based on them. In the evaluation step, we train a neural network from scratch on the distilled dataset and assess its accuracy and robustness over 500 epochs with a learning rate of 0.01. To ensure fairness and reliability, we train five identical models under consistent conditions, computing the mean and variance of their accuracy and robustness as our final reported results.

A key aspect of our setup is image preprocessing. To evaluate robustness, we normalize pixel values to [0,1]. Unlike prior studies~\cite{cazenavette2022dataset, du2023minimizing, liu2024dataset}, we avoid additional normalization steps such as ZCA~\cite{krizhevsky2009learning} to maintain consistency in robustness evaluations. This may cause slight discrepancies between our results and originally reported baseline performances. We obtained dataset distillation results from open-source code and conducted additional experiments.

We compared our method against a comprehensive set of existing algorithms to demonstrate the effectiveness of our algorithm in improving robustness, including Dataset Condensation (DC)~\cite{zhao2020dataset}, Differentiable Siamese Augmentation (DSA) \cite{zhao2021dataset}, gradient-matching methods Distribution Matching (DM) \cite{zhao2023dataset},  trajectory matching method (MTT) \cite{cazenavette2022dataset}, Flat Trajectory Distillation (FTD) \cite{du2023minimizing} and Automatic Training Trajectories (ATT) \cite{liu2024dataset}. 

\noindent\textbf{Model Architecture.} To maintain consistency with the existing literature, we employ a straightforward convolutional neural network with depth = 3 (ConvNetD3) \cite{gidaris2018dynamic} as our test model(ConvNetD4 for Tiny ImageNet). The architecture comprises 3 repeated blocks, each containing the following layers in sequence: a 3×3 convolutional layer with 128 output channels, an instance normalization layer, a ReLU activation layer, and a 2×2 average pooling layer. After stacking multiple convolutional blocks, the network concludes with a fully connected layer.

\begin{table*}[ht]
\centering
\caption{Performance comparison of standard and adversarial accuracy (\%) on CIFAR-100 across different methods with varying IPC (Images Per Class) values. We used PGD-10 with $\epsilon$ = 4/255 to test models' robustness. We reproduced all baseline results in the adversarial defense scenario. The \textbf{best} results in each column are highlighted for clarity.}
{%
\small
\begin{tabular}{l c c c c c c}
    \toprule
    IPC & \multicolumn{2}{c}{1} & \multicolumn{2}{c}{10} & \multicolumn{2}{c}{50} \\
    \cmidrule(lr){2-3} \cmidrule(lr){4-5} \cmidrule(lr){6-7}
    Performance & Standard acc & Adversarial acc & Standard acc & Adversarial acc & Standard acc & Adversarial acc \\
    \midrule
    DC~\cite{zhao2020dataset} & 11.56$\pm$0.35 & 2.98$\pm$0.10 & 24.55$\pm$0.05 & 4.04$\pm$0.18 & - & - \\
    DM~\cite{zhao2023dataset} & - & - & 28.51$\pm$0.29 & 1.81$\pm$0.11 & 34.47$\pm$0.25 & 0.57$\pm$0.04 \\
    DSA~\cite{zhao2021dataset} & 11.30$\pm$0.30 & 2.88$\pm$0.15 & 26.72$\pm$0.19 & 3.58$\pm$0.25 & - & - \\ 
    MTT~\cite{cazenavette2022dataset} & 14.54$\pm$0.66 & 2.29$\pm$0.19 & 29.32$\pm$0.69 & 1.50$\pm$0.29 & 38.35$\pm$0.61 & 0.94$\pm$0.10 \\
    FTD~\cite{du2023minimizing} & 8.34$\pm$0.25 & 2.32$\pm$0.16 & 26.61$\pm$0.81 & 1.17$\pm$0.16 & 38.03$\pm$0.32 & 0.82$\pm$0.06 \\
    ATT~\cite{liu2024dataset} & 14.57$\pm$0.40 & 2.68$\pm$0.18 & 29.50$\pm$0.30 & 2.26$\pm$0.14 & 35.09$\pm$0.10 & 1.31$\pm$0.11 \\
    \midrule
    MAT (Ours) & \textbf{17.58$\pm$0.38} & \textbf{4.11$\pm$0.19} & \textbf{34.21$\pm$0.32} & \textbf{4.53$\pm$0.29} & \textbf{39.52$\pm$0.27} & \textbf{3.98$\pm$0.16} \\
    \bottomrule
\end{tabular}%
}
\label{tab:cifar100}
\end{table*}

\begin{table}[ht]
\centering
\caption{Performance comparison of standard and adversarial accuracy (\%) on Tiny ImageNet across different methods with 10 or 50 images per class. We used PGD-10 with $\epsilon$ = 2/255 to test models' robustness. We reproduced all baseline results in the adversarial defense scenario. The \textbf{best} results in each column are highlighted for clarity.}
\resizebox{\columnwidth}{!}
{%
\begin{tabular}{l c c c c}
    \toprule
    IPC & \multicolumn{2}{c}{10} & \multicolumn{2}{c}{50} \\
    \cmidrule(lr){2-3} \cmidrule(lr){4-5} 
    Performance & Standard acc & Adversarial acc & Standard acc  & Adversarial acc \\
    \midrule
    DM~\cite{zhao2023dataset} & 12.37$\pm$0.24 & 0.04$\pm$0.01 & 20.29$\pm$0.32 & 0.00$\pm$0.00 \\
    MTT~\cite{cazenavette2022dataset} & \textbf{14.80$\pm$0.29} & 0.00$\pm$0.00 & 21.04$\pm$0.27 & 0.61$\pm$0.03  \\
    ATT~\cite{liu2024dataset} & 11.56$\pm$0.31 & 0.01$\pm$0.00 & \textbf{23.49$\pm$0.24} & 0.98$\pm$0.07  \\
    \midrule
    MAT (Ours) & 12.22$\pm$0.35 & \textbf{1.67$\pm$0.09} & 22.06$\pm$0.14 & \textbf{4.21$\pm$0.64}  \\
    \bottomrule
\end{tabular}%
}
\label{tab:tiny}
\end{table}

\subsection{Results} 

\textbf{CIFAR-10 and CIFAR-100~\cite{krizhevsky2009learning}.} As can be seen from the results in the Tables \ref{tab:cifar10} and \ref{tab:cifar100}, compared to previous methods, our method embeds robustness into the data with comparable accuracy, which allows the model to rely on natural training alone to obtain robustness.

\noindent\textbf{Tiny ImageNet~\cite{deng2009imagenet}.} Compared with CIFAR, Tiny ImageNet is more difficult and takes longer to reproduce the results, so we only reproduce and compare three most typical trajectory matching-based algorithms. We set attack to PGD-10 with $\epsilon$ = 2/255 during test and result is shown in Table \ref{tab:tiny}. From the table, it can be seen that obtaining robustness on Tiny ImageNet is much more difficult, but our algorithm still has significant improvements compared to existing algorithms.

\noindent\textbf{Performance on various attacks.} It is crucial for an algorithm to withstand various adversarial attacks, so we evaluated the model trained on the distilled dataset generated by our algorithm under various attacks to ensure broad-spectrum robustness. Specifically, we generated a distilled dataset using the ConvNetD3 architecture in CIFAR-10 with IPC = 1 and then evaluated the robustness of the model against projected gradient descent (PGD)~\cite{mkadry2017towards}, fast gradient sign method (FGSM)~\cite{goodfellow2014explaining} and AutoAttack~\cite{croce2020reliable}. We retained the same hyperparameters described earlier and set the adversarial perturbation magnitude ($\epsilon$) to 4/255 for validation.

The results in Table~\ref{tab:V/A} indicate that the model exhibits robust defensive capabilities in all adversarial attacks. Specifically, our method consistently outperforms the baseline in mitigating the impact of various attacks, demonstrating its effectiveness in enhancing model robustness. These findings confirm that our approach not only provides resilience against specific adversarial strategies but also ensures comprehensive protection across a diverse range of attack methodologies. Overall, the result underscores the efficacy of our algorithm in achieving robust adversarial defense, thereby validating its potential for practical applications where security and reliability are paramount.

\begin{table}[!t]
    \centering
    \caption{Adversarial accuracy (\%) performance comparison on CIFAR-10 across various attacks with an 10 images per class. Standard means performance on the standard test set (without attack). The strength of attack($\epsilon$) is set to 4/255. The \textbf{best} results in each column are highlighted for clarity.}
    \setlength{\tabcolsep}{3pt}
    \resizebox{\columnwidth}{!}{%
    \begin{tabular}{c|cccc}
        \toprule
         \multicolumn{1}{c|}{Method} & \multicolumn{1}{c}{Standard} & \multicolumn{1}{c}{PGD~\cite{mkadry2017towards}} & \multicolumn{1}{c}{AutoAttack~\cite{croce2020reliable}} & \multicolumn{1}{c}{FGSM~\cite{goodfellow2014explaining}} \\
         \midrule
        ATT~\cite{cazenavette2022dataset} & \textbf{38.61$\pm$1.01} & 6.48$\pm$0.94 & 8.21$\pm$0.96 &  14.20$\pm$0.77\\
        MAT (ours) & 35.96$\pm$0.74 & \textbf{14.33$\pm$0.63} & \textbf{12.21$\pm$0.06} & \textbf{16.08$\pm$0.74} \\
        \bottomrule
    \end{tabular}%
    }
    \label{tab:V/A}
\end{table}

\begin{table*}[ht]
    \centering
    \caption{Performance comparison of cross-architecture generalization between MAT and MTT. The table reports both standard and adversarial accuracy (\%) for each architecture. The results of this experiment are based on CIFAR-10 with 1 image per class. The \textbf{best} results in each column are highlighted for clarity.}
    \resizebox{\textwidth}{!}{%
    \begin{tabular}{c|cccccccc}
        \toprule
         \multicolumn{1}{c|}{Method} & \multicolumn{2}{c}{ConvNetD3~\cite{gidaris2018dynamic}} & \multicolumn{2}{c}{AlexNet~\cite{krizhevsky2012imagenet}} & \multicolumn{2}{c}{VGG11~\cite{simonyan2014very}} & \multicolumn{2}{c}{ResNet18~\cite{he2016deep}} \\
        \cmidrule(lr){1-1}\cmidrule(lr){2-3} \cmidrule(lr){4-5} \cmidrule(lr){6-7}\cmidrule(lr){8-9}
        \multicolumn{1}{c|}{Performance} & Standard acc & Adversarial acc & Standard acc & Adversarial acc & Standard acc & Adversarial acc & Standard acc & Adversarial acc \\
        \midrule
        \multirow{1}{*}{MTT \cite{cazenavette2022dataset}} & \textbf{38.61$\pm$1.01} & 6.48$\pm$ 0.94 & 17.15$\pm$3.80 & 7.54$\pm$1.92 & 25.37$\pm$1.14 & 2.28$\pm$0.28 & 28.24$\pm$0.74 & 2.07$\pm$0.39 \\
        \multirow{1}{*}{MAT (ours)} & 35.96$\pm$0.74 & \textbf{14.33$\pm$0.63} & \textbf{19.31$\pm$3.15} & \textbf{11.75$\pm$1.88} & \textbf{29.01$\pm$1.60} & \textbf{5.16$\pm$1.85} & \textbf{30.68$\pm$1.12} & \textbf{7.06$\pm$0.80} \\
        \bottomrule
    \end{tabular}%
}
    \label{tab:C/A}
\end{table*}

\begin{table}[!t]
    \centering
    \caption{Standard and adversarial accuracy (\%) performance comparison of different adversarial training methods used by our method when generating expert trajectories, including the natural training (without using adversarial training), PGD-AT, TRADES, and MART. We set IPC = 1 on the CIFAR-10 to distill dataset and get these experiment results.}
    {\small
    
    \begin{tabular}{c|cc}
        \toprule
         \multicolumn{1}{c|}{Adversarial training method} & \multicolumn{1}{c}{Standard acc} & \multicolumn{1}{c}{Adversarial acc} \\
         \midrule
        Natural & \textbf{37.30$\pm$0.69} & 5.58$\pm$1.13 \\
        PGD-AT~\cite{mkadry2017towards} & 35.96$\pm$0.74 & \textbf{14.33$\pm$0.63}\\
        TRADES~\cite{zhang2019theoretically} & 36.09$\pm$9.87 & 7.27$\pm$0.78 \\
        MART~\cite{wang2019improving} & 35.50$\pm$0.83 & 9.99$\pm$1.32 \\
        \bottomrule
    \end{tabular}%
    }
    \label{tab:AT-method}
\end{table}
\noindent\textbf{Cross-architecture generalization.} Measuring the availability of a dataset across different architectures, cross-architecture generalization is a crucial metric in dataset distillation. In this section, we evaluate the cross-architecture generalization capabilities of our robust distilled dataset. Consistent with prior methods, we trained a robust distilled dataset with IPC = 1 on the CIFAR-10 dataset using a ConvNetD3 \cite{gidaris2018dynamic} architecture. Building upon this, we tested the cross-architecture robustness of our method on several classic neural network architectures, including ResNet-18 \cite{he2016deep}, VGG-11 \cite{simonyan2014very}, AlexNet \cite{krizhevsky2012imagenet}, and ConvNetD3~\cite{gidaris2018dynamic}.


During the evaluation of this metric, we maintained consistency in our experimental hyper-parameters, as previously described. For comparison, we selected the MTT \cite{cazenavette2022dataset}, the most representative trajectory matching-based method as our baseline.
As shown in Table \ref{tab:C/A}, the results indicate that architectural changes introduce some performance degradation, however, the models still retain a significant degree of robustness. Notably, compared to the baseline MTT method, our approach consistently outperforms MTT across different architectures. This demonstrates that our method not only preserves robustness when transferring to various network architectures but also maintains a leading performance position relative to established baseline methods. These findings underscore the effectiveness of the approach in achieving cross-architecture generalization, thereby enhancing the practical utility of distilled datasets in diverse application environments.


\noindent\textbf{Different adversarial trajectories.} To demonstrate the versatility of our method in fitting trajectories derived from various adversarial training strategies, we conducted an ablation study focusing on different adversarial training trajectories. Specifically, we selected several classical adversarial training approaches, including Adversarial Training (AT) \cite{mkadry2017towards}, TRADES \cite{zhang2019theoretically}, and MART \cite{wang2019improving}, to generate robust expert trajectories. We subsequently generated new distilled datasets tailored to each adversarial training method. The ablation experiments were designed to assess whether our method can be effectively adapted to any trajectory imparted by different adversarial training techniques. We aimed to verify that our dataset distillation process is broadly applicable across multiple robust training paradigms.
The experimental results, summarized in Table \ref{tab:AT-method}, indicate that our method successfully fits the trajectories generated by each of the selected adversarial training methods. Across AT \cite{mkadry2017towards}, TRADES \cite{zhang2019theoretically} and MART \cite{wang2019improving}, our approach consistently produced distilled datasets that preserved better robustness than the natural method. These findings confirm that our method is capable of accommodating various adversarial training methodologies, thereby enhancing its applicability and effectiveness in different adversarial defense scenarios.


\begin{table}[!t]
    \centering
    \caption{Ablation study results of the impact of adversarial training (AT) and Exponential Moving Average (EMA) on performance. The table reports standard and adversarial accuracy (\%), comparing the use of adversarial training and the EMA mechanism when generating expert trajectories. This comparison is based on CIFAR-10 (IPC = 10) with attack PGD-10 ($\epsilon$ = 4/255).}
    {\small
    \begin{tabular}{cc|cc}
        \toprule
        AT & EMA & Standard acc & Adversarial acc \\
        \midrule
         \color{red}\XSolidBrush & \color{red}\XSolidBrush & \textbf{55.28$\pm$0.87} & 5.55$\pm$0.39 \\
        \color{blue}\Checkmark & \color{red}\XSolidBrush & 50.62$\pm$0.60 & 4.80$\pm$0.51 \\
         \color{red}\XSolidBrush & \color{blue}\Checkmark & 51.65$\pm$0.55 & 7.03$\pm$0.80 \\
        \color{blue}\Checkmark & \color{blue}\Checkmark & 53.03$\pm$0.71 & \textbf{11.85$\pm$0.80}\\
        \bottomrule
    \end{tabular}
    }
    \label{tab:ablation-modules}
\end{table}

\subsection{Ablation study}
\begin{figure}[t!]
    \centering
    \includegraphics[width=1\columnwidth]{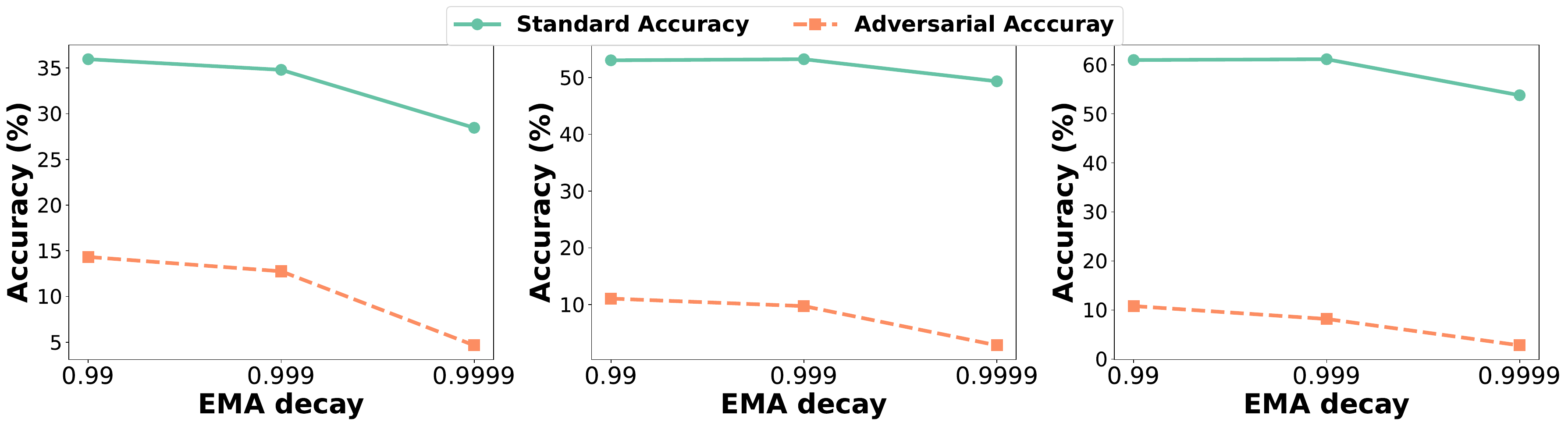}

    \begin{subfigure}[b]{0.32\columnwidth}
        \centering
        \caption{IPC 1}
        \label{fig:decay_ipc1}
    \end{subfigure}
    \begin{subfigure}[b]{0.32\columnwidth}
        \centering
        \caption{IPC 10}
        \label{fig:decay_ipc10}
    \end{subfigure}
    \begin{subfigure}[b]{0.32\columnwidth}
        \centering
        \caption{IPC 50}
        \label{fig:decay_ipc50}
    \end{subfigure}
    \caption{Impact of hyper-parameters EMA decay on model performance. Based on CIFAR-10, we set the EMA decay rate to 0.99, 0.999, and 0.9999 while keeping other hyper-parameters fixed and observed the effect on performance.}
    \label{fig:ema}
\end{figure}

\textbf{Adversarial training and EMA.} Our method employs two key modules that distinguish it from prior dataset distillation approaches: considering adversarial training trajectories and strategically applying the EMA mechanism to achieve smoother optimization. To evaluate their effectiveness, we conducted ablation studies on CIFAR-10 with IPC = 10. As shown in Table \ref{tab:ablation-modules}, both modules significantly enhance the distilled dataset’s performance. Specifically, without the EMA mechanism, adversarial training leads to rapidly fluctuating trajectories, which hinder the dataset’s ability to fit expert trajectories, resulting in severe performance degradation. Conversely, removing adversarial training entirely prevents the distilled dataset from acquiring adversarial robustness, making it ineffective in improving model resilience against adversarial attacks.

\noindent\textbf{Hyper-parameters.} For the hyperparameter EMA decay rate, we tested values of 0.99, 0.999, and 0.9999 to evaluate their impact on the performance of the model trained on distilled dataset. All other parameters were kept the same as described earlier. The results, presented in Fig. \ref{fig:ema}, reveal that transitioning the EMA decay rate from 0.99 to 0.999 did not result in significant variations in accuracy or robustness metrics. However, a decay rate of 0.9999 will degrade the performance of the expert trajectory, thereby affecting the distilled data. We consider this is due to the excessive decay rate affecting the learning of expert trajectories.

\begin{figure}[t!]
    \centering
    \includegraphics[width=0.9\columnwidth]{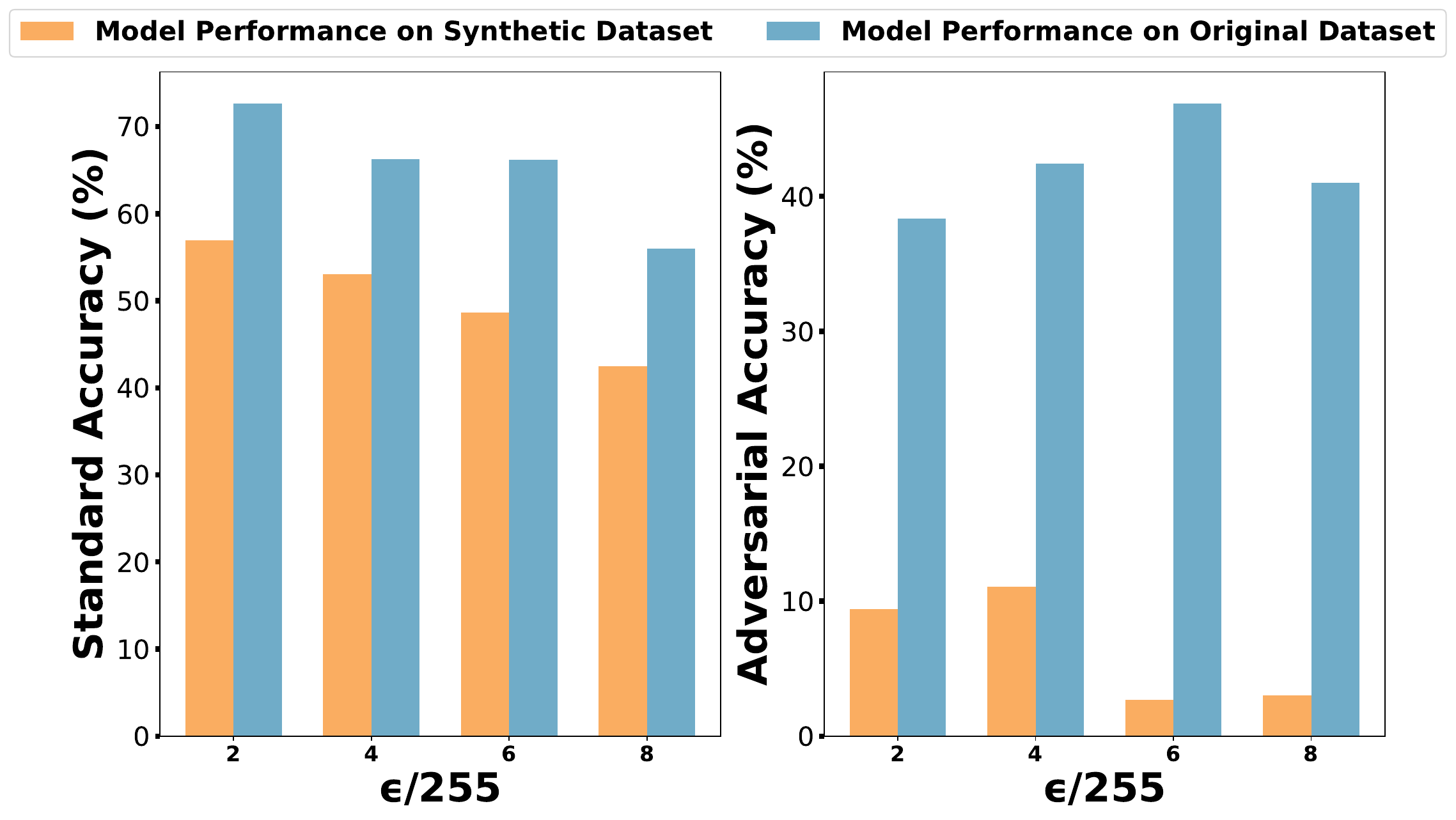}

    \begin{subfigure}[b]{0.42\columnwidth}
        \centering
        \caption{Standard Accuracy}
        \label{fig:standard_accuracy}
    \end{subfigure}
    \begin{subfigure}[b]{0.42\columnwidth}
        \centering
        \caption{Adversarial Accuracy}
        \label{fig:adv_accuracy}
    \end{subfigure}
    
    \caption{Influence on $\epsilon$ on model performance. We compare both the standard and adversarial accuracy of synthetic data (IPC=10) with the original CIFAR-10 data. We set $\epsilon$ of attack when generating the buffers to 2/255, 4/255, 6/255, and 8/255. All other parameters were kept constant. }
    \label{fig:eps}
\end{figure}

For the hyper-parameter strength of the adversarial attack, we set $\epsilon$ of attack to 2/255, 4/255, 6/255, and 8/255. All other parameters were kept constant throughout these experiments. During the evaluation phase, we uniformly applied an adversarial perturbation of $\epsilon$ = 4/255 across all tests to maintain consistency. As shown in Fig.~\ref{fig:eps}, increasing the attack strength during adversarial training adversely affected the performance. Higher perturbation magnitudes (6/255 and 8/255) led to a noticeable decline in both accuracy and robustness. We hypothesize that this degradation is attributable to the substantial influence of expert trajectories in the dataset distillation process. Introducing excessively strong adversarial attacks during training may compromise the integrity of these trajectories, thereby diminishing the effectiveness of the distilled dataset.

\subsection{Further analysis}
\begin{figure}[!t]
    \centering
    \begin{subfigure}[b]{0.32\columnwidth}
        \centering
        \includegraphics[width=\columnwidth]{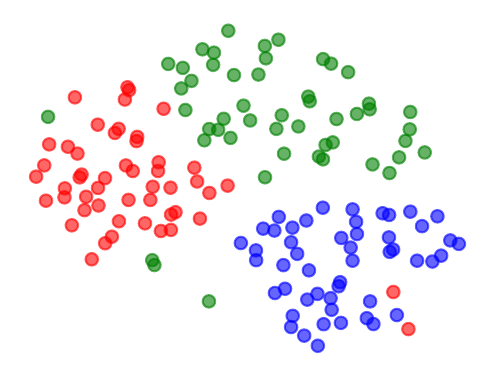}
        \label{fig:sub1}
        \caption{MTT~\cite{cazenavette2022dataset}}
    \end{subfigure}
    \begin{subfigure}[b]{0.32\columnwidth}
        \centering
        \includegraphics[width=\columnwidth]{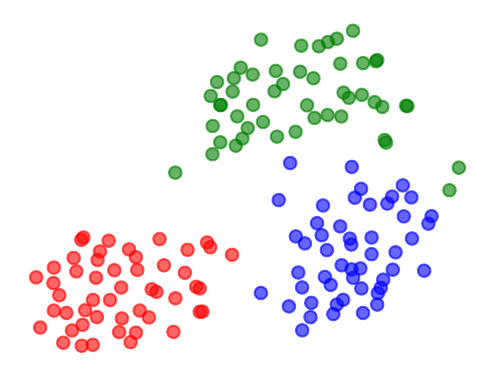}
        \label{fig:sub2}
        \caption{FTD~\cite{du2023minimizing}}
    \end{subfigure}
    \begin{subfigure}[b]{0.32\columnwidth}
        \centering
        \includegraphics[width=\columnwidth]{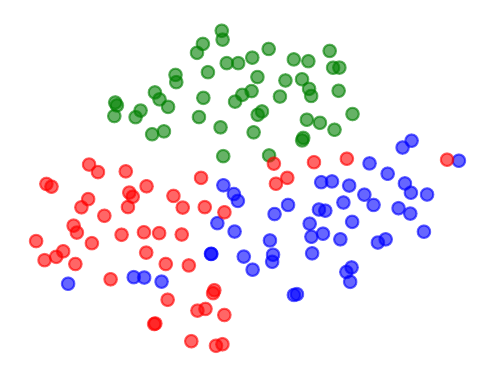}
        \label{fig:sub3}
        \caption{MAT (ours)}
    \end{subfigure}
    \caption{Feature difference between robust dataset and normal dataset. We used a model fully trained on the original CIFAR-10 dataset as a feature extractor to extract features of datasets generated by MTT~\cite{cazenavette2022dataset}, FTD~\cite{du2023minimizing}, and our MAT. For each method, we extract features from the first three classes and visualize them using t-SNE~\cite{van2008visualizing}.} 
    \label{fig:three_images}
\end{figure}

\textbf{Differences in feature distribution.} As shown in Fig.~\ref{fig:three_images}, we compare features of distilled datasets generated by our method with those from existing approaches on CIFAR-10 with setting 50 images per class. We used a ConvNetD3~\cite{gidaris2018dynamic} model fully trained on the original CIFAR-10 dataset with its classifier removed to function as a feature extractor. By extracting features from all samples of the first three classes in each distilled dataset and applying t-SNE~\cite{van2008visualizing} for visualization, we observed that the feature extractor retains high discriminative power for datasets produced by other methods. However, it fails to effectively distinguish our robust distilled dataset, similar to its behavior when confronted with adversarial samples. These findings demonstrate that our distilled dataset successfully emulates adversarial examples, enabling models to acquire inherent adversarial robustness with natural training.

\section{Conclusion}
\label{sec:discuss}

In this paper, we try to address Robust Dataset Distillation, an important but overlooked problem, by introducing a smooth adversarial trajectory matching approach. This method enables models trained on the distilled dataset to achieve adversarial robustness without explicit adversarial training. Our framework incorporates two key modules over previous work: integrating adversarial training into trajectory generation and adopting EMA mechanism to smooth expert trajectories. Through comprehensive experiments, we have demonstrated that models trained on our distilled datasets attain improved adversarial robustness, underscoring the practicality and potential of robust dataset distillation for future research. This is an exploration step, and we look forward to more research along this direction.


\section{Acknowledgements}

This work is supported in part by the Young Elite Scientists Sponsorship Program by CAST (2023QNRC001), National Natural Science Foundation of China (62192783, 62276128),  and Jiangsu Natural Science Foundation (BK20221441).
\clearpage
{
    \small
    \bibliographystyle{ieeenat_fullname}
    \bibliography{main}
}


\end{document}